\documentclass[11pt]{article}
\PassOptionsToPackage{hyphens}{url}
\usepackage[final]{acl}
\usepackage{times}
\usepackage{latexsym}
\usepackage[T1]{fontenc}
\usepackage[utf8]{inputenc}
\usepackage{microtype}
\usepackage{graphicx}
\usepackage{booktabs}
\usepackage{amsmath}
\usepackage{multirow}
\usepackage{enumitem}
\usepackage{caption}
\usepackage{subcaption}
\usepackage{float}
\newcommand{\pp}{\,\text{pp}}

\title{Cross-Lingual Gaps in Visual STEM Reasoning:\\Auditing VLMs Across Six Indian Languages}

\author{Swastik R \\
  Indian Institute of Information Technology, Raichur \\
  \texttt{cs23b1075@iiitr.ac.in}}

\begin{document}
\maketitle

\begin{abstract}
Vision-language models (VLMs) are benchmarked on diagram-grounded science, technology, engineering, and mathematics (STEM) reasoning mostly in English, so it is unclear how much their accuracy changes when the same problem is posed in another language.
We audit this for six Indian languages: 980 questions from MathVista, ScienceQA, and MMMU are machine-translated into Hindi, Bengali, Marathi, Tamil, Telugu, and Kannada with IndicTrans2 while the images stay unchanged.
We evaluate seven open-weight VLMs (Qwen2.5-VL-7B/32B, Qwen3-VL-30B, InternVL2.5-8B, Aya-Vision-8B, Gemma~3-27B, Llama-4-Maverick) and GPT-4o, plus no-image and chain-of-thought (CoT) ablations on Qwen2.5-VL-7B (68{,}600 inference records).
Accuracy falls 9.8--24.9 percentage points (pp) from English to the Indian-language mean, and every model loses more on the Dravidian than on the Indo-Aryan languages (by 0.1--13.2$\pp$).
For Qwen2.5-VL-7B with our prompt, CoT lowers Bengali ($-$14.4$\pp$) and Kannada ($-$11.4$\pp$) accuracy while barely changing English, but these losses coincide with more answer-extraction failures.
Neither Aya-Vision-8B, trained for 23 languages of which only Hindi is among ours, nor scaling Qwen2.5-VL from 7B to 32B closes the gap, and response-level diagnostics show that fluent target-language output and correct answers can come apart.
We quantify how machine translation, answer extraction, and English text in the images bound these measurements.
Data and code are publicly available.\footnote{Data: \url{https://huggingface.co/datasets/Swastikr/multilingual-vlm-reasoning}; code: \url{https://github.com/QuantumByte-01/multilingual-vlm-reasoning-audit}.}
\end{abstract}

\section{Introduction}
\label{sec:intro}

India's National Education Policy~2020 asks that, wherever possible, the medium of instruction until at least Grade~5 be the home language, mother tongue, or regional language, and calls for bilingual science and mathematics materials~\cite{nep2020}.
If VLMs are ever to support STEM learning in these languages, a prior measurement question must be answered: on the \emph{same} diagram-grounded problem, how much does a model's accuracy change when the question is posed in an Indian language instead of English?
We treat this strictly as pre-deployment evaluation. Even in English, the models we test answer only 42--65\% of our questions correctly, far from what tutoring would require; our object of study is therefore the \emph{relative} disparity between languages, which any future system built on these models would inherit.

Existing Indian-language VLM benchmarks do not answer this question.
They probe cultural understanding, factual visual question answering (VQA), or broad optical character recognition (OCR) proficiency~\cite{faraz2025indicvisionbench,maji2025drishtikon,surana2026viraasat,khan2025chitrarth}, but none isolate mathematical or scientific reasoning with a controlled design.
Multilingual vision-language work outside India (MaRVL, xGQA, VLURes, CVQA, ALM-Bench, Pangea) rarely combines Indian languages with math and science reasoning suites~\cite{liu2021marvl,pfeiffer2022xgqa,romero2024cvqa,atuhurra2025vlures,vayani2024alm,yue2024pangea}, and India-focused VLM architectures such as Chitrarth~\cite{khan2025chitrarth} have no reasoning-specific evaluation to test against.
We hold images fixed and vary only the question text across six Indian languages spanning two families and five scripts. This removes visual content as an explanation for differences on the same question, but the manipulated variable is the whole text condition (the language, its machine translation, its tokenization, and the answer format a model produces); Section~\ref{sec:scope} states what the design does and does not isolate.

\begin{figure}[t]
\centering
\includegraphics[width=\columnwidth]{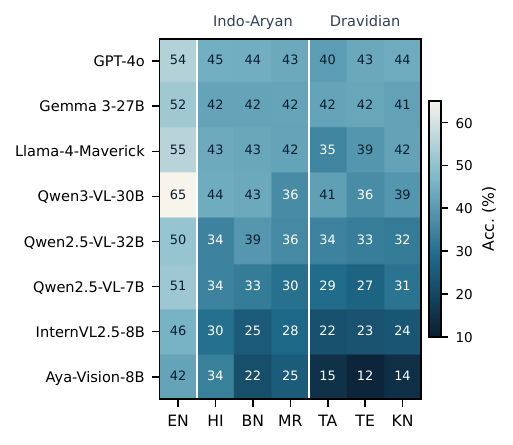}
\caption{Every model is less accurate in every Indian language than in English, and averaged over models each Dravidian language (TA/TE/KN) loses more than each Indo-Aryan language (HI/BN/MR). Cells show accuracy (\%) on the same 980 image-fixed questions; white lines separate English, Indo-Aryan, and Dravidian columns.}
\label{fig:heatmap}
\end{figure}

We evaluate eight VLMs, seven open-weight (Qwen2.5-VL-7B and -32B, Qwen3-VL-30B, InternVL2.5-8B, Aya-Vision-8B, Gemma~3-27B, Llama-4-Maverick) and one proprietary (GPT-4o), on translated MathVista~\cite{lu2024mathvista}, ScienceQA~\cite{lu2022scienceqa}, and MMMU~\cite{yue2024mmmu} questions, together with no-image and CoT ablations on Qwen2.5-VL-7B, for a total of 68{,}600 inference records.
Figure~\ref{fig:heatmap} summarizes the pattern: every model is less accurate in every Indian language than in English, and averaged over models each Dravidian language loses more than each Indo-Aryan language.

\textbf{Findings.}
\begin{itemize}[nosep,leftmargin=1.3em]
\item Gemma~3-27B shows the smallest English-to-Indian drop (9.8$\pp$); GPT-4o has the highest Indian-language mean accuracy (43.2\%); the 7--8B open models lose 20.4--21.9$\pp$.
\item The Dravidian drop exceeds the Indo-Aryan drop for all eight models, from 0.1$\pp$ (Gemma~3-27B) to 13.2$\pp$ (Aya-Vision-8B, whose declared languages include Hindi but no Dravidian language).
\item For Qwen2.5-VL-7B with a translated ``think step by step'' instruction, CoT lowers Bengali and Kannada accuracy by 14.4 and 11.4$\pp$ while barely affecting English (Figure~\ref{fig:ablation}); the affected languages also show 13--17$\pp$ more extraction failures, so this is a single-model, single-prompt finding.
\item Removing images costs the same model 15.5$\pp$ in English but 4.9--9.7$\pp$ in Indian languages (Figure~\ref{fig:ablation}); under-used vision when the text is harder to parse is one hypothesis, but the English no-image condition also produces many more refusal-like responses (Section~\ref{sec:ablations}).
\item Scaling Qwen2.5-VL from 7B to 32B gains 4.3$\pp$ on Indian languages while losing 1.0$\pp$ on English, narrowing but not closing the gap.
\item Response-length, script-confusion, and English-token-leak patterns (Figures~\ref{fig:leak}, \ref{fig:confusion}, and~\ref{fig:length}) differ across models with similar accuracy, so fluent target-language output and correct answers can come apart.
\end{itemize}

\section{Related Work}
\label{sec:related}

\paragraph{Indian multimodal evaluation.}
IndicVisionBench~\cite{faraz2025indicvisionbench} covers ten Indian languages with over 37K QA pairs but targets cultural understanding and OCR.
DRISHTIKON~\cite{maji2025drishtikon} and VIRAASAT~\cite{surana2026viraasat} evaluate cultural knowledge and geographic diversity.
Chitrarth~\cite{khan2025chitrarth} introduces an India-focused VLM architecture and a BharatBench evaluation suite spanning speech, OCR, and embeddings, but no reasoning-specific evaluation.
None of these isolate STEM reasoning with a fixed-image, language-only control.

\paragraph{Multilingual VLM evaluation beyond India.}
MaRVL~\cite{liu2021marvl} tests visually grounded reasoning in five languages, including Tamil, but relies on culturally situated images and predates current VLMs.
xGQA~\cite{pfeiffer2022xgqa} extends GQA to seven languages, of which only Bengali is Indian and without reasoning-focused content; VLURes~\cite{atuhurra2025vlures} covers four languages including Urdu.
ALM-Bench~\cite{vayani2024alm} and CVQA~\cite{romero2024cvqa} include partial Indian-language coverage but do not focus on reasoning.
Pangea~\cite{yue2024pangea}, a 39-language multimodal large language model (LLM), is the closest open multilingual effort, yet it too lacks a reasoning-focused evaluation suite.

\paragraph{Visual reasoning benchmarks.}
MathVista~\cite{lu2024mathvista} evaluates mathematical reasoning in visual contexts, and MATH-Vision~\cite{wang2024mathvision} extends this to competition-level problems.
ScienceQA~\cite{lu2022scienceqa} covers K-12 science with diagrams; MMMU~\cite{yue2024mmmu} targets college-level multimodal reasoning across 30 subjects, of which we use all 180 validation questions of six STEM subjects.
All three are English-only; we extend them cross-lingually with paired, image-fixed translations.

\paragraph{Multilingual reasoning.}
\citet{shi2023mgsm} show that English chain-of-thought outperforms native-language CoT on grade-school math across ten languages, without images.
\citet{ahuja2023mega} broaden multilingual generative evaluation across tasks and languages, and \citet{saji2025reasoning} find that language reasoning models default to English reasoning chains regardless of prompt language on text-only benchmarks.
We extend this line of inquiry to the visual domain, where images may partly compensate for weak text comprehension.

\paragraph{Indian NLP resources.}
IndicTrans2~\cite{gala2023indictrans2} provides open neural machine translation for all 22 scheduled Indian languages and outperforms Google Translate and Meta's No Language Left Behind (NLLB-54B); we use it as our primary translation system, complemented by the broader Indic natural language processing (NLP) infrastructure described by~\citet{doddapaneni2023towards}.
We chose it because it is openly licensed (MIT), covers all 22 scheduled languages, and runs locally and deterministically; re-translating with newer LLM-based systems such as Sarvam-Translate is left to future work.

\section{Method}
\label{sec:method}

\subsection{Dataset Construction}
We assemble 980 image-grounded questions from three English benchmarks. From each source we take the first eligible items in the official dataset order, with no random sampling or stratification (Appendix Table~\ref{tab:composition}).
\textbf{MathVista}~\cite{lu2024mathvista}: the first 400 testmini questions (IDs 1--466) whose text contains at least four purely alphabetic tokens; this translatability filter skips 66 questions that are very short, mostly symbolic, or Chinese (33 of the 66), and 7 IQ-test questions remain (195 multiple-choice, 205 free-form numeric).
\textbf{ScienceQA}~\cite{lu2022scienceqa}: the first 400 of the 1{,}209 natural-science questions in the image test split (physics 133, biology 125, science practices 56, earth science 51, chemistry 32, other 3; grades 2--8), all multiple-choice; the optional textual context (present for 376 of the 400) is not given to the models.
\textbf{MMMU}~\cite{yue2024mmmu}: all 180 validation questions of six STEM subjects (Math, Physics, Chemistry, Architecture and Engineering, Mechanical Engineering, Computer Science; 167 multiple-choice, 13 free-form).
Each question keeps the same image in all seven languages; only the text changes.
Text inside images is \emph{not} translated: by OCR, 43.2\% of items show English words in the image (axis titles, legends, labels; 5.1\% have tables or running text), while 56.8\% contain no text or only language-neutral symbols such as digits, variables, and units (Appendix~\ref{app:imagetext}). The measured gap is therefore the effect of question language \emph{given an English-annotated image}.
We use subsets because every item is run in seven languages by eight models plus two ablations (6{,}860 calls per model run, 68{,}600 in total). With 980 items, 95\% bootstrap intervals of per-language accuracies have half-widths of 1.9--3.3\pp{} (Appendix~\ref{app:ci}); per-benchmark results rest on 400, 400, and 180 items and are less precise (about $\pm$7\pp{} on MMMU). All results describe these subsets, not the full benchmarks.

\subsection{Languages}
Table~\ref{tab:languages} lists the six Indian languages plus English.
They are the five scheduled languages with the most mother-tongue speakers in the 2011 Census (Hindi, Bengali, Marathi, Telugu, Tamil) plus Kannada (eighth), together returned as mother tongue by 902 million people, or 74.5\% of the population~\cite{census2011lang}.
They cover the two largest language groups of India (Indo-Aryan, 78.1\%, and Dravidian, 19.6\% of the population by mother tongue) with three languages each and five scripts, and all are supported by IndicTrans2.
Including Kannada gives a balanced design with three Indo-Aryan and three Dravidian languages.
Hindi and Marathi share Devanagari, which lets us compare two languages without a script difference; with a single same-script pair this comparison is suggestive rather than a controlled test of script effects.

\begin{table}[t]
\centering
\small
\setlength{\tabcolsep}{2.5pt}
\caption{Languages studied. The six Indian languages are returned as mother tongue by 74.5\% of India's population and cover both major families and five scripts, with one same-script pair (Hindi/Marathi). Family and Spk.\ (millions, mother tongue) follow the 2011 Census~\cite{census2011lang}; Germanic and Indo-Aryan are branches of Indo-European. Jacc.\ = set-based character Jaccard and chrF = sentence chrF between IndicTrans2 and an independent Gemini~2.0 Flash forward translation of the question (50 questions per language; MT-to-MT, automatic); unrelated question pairs score .45--.52 (Jacc.) and 11.7--16.2 (chrF). Takeaway: both proxies are high and do not follow the family pattern, but neither measures adequacy (Section~\ref{sec:translation}).}
\label{tab:languages}
\begin{tabular}{@{}lllrcc@{}}
\toprule
\textbf{Language} & \textbf{Script} & \textbf{Family} & \textbf{Spk.} & \textbf{Jacc.} & \textbf{chrF} \\
\midrule
English (en) & Latin & Germanic & --- & --- & --- \\
Hindi (hi) & Devanagari & Indo-Aryan & 528.3 & .844 & 68.1 \\
Bengali (bn) & Bengali & Indo-Aryan & 97.2 & .806 & 62.9 \\
Marathi (mr) & Devanagari & Indo-Aryan & 83.0 & .791 & 59.2 \\
Tamil (ta) & Tamil & Dravidian & 69.0 & .811 & 65.1 \\
Telugu (te) & Telugu & Dravidian & 81.1 & .803 & 62.6 \\
Kannada (kn) & Kannada & Dravidian & 43.7 & .791 & 60.1 \\
\bottomrule
\end{tabular}
\end{table}

\subsection{Translation}
\label{sec:translation}
IndicTrans2~\cite{gala2023indictrans2} (en--indic 1B, beam size 4) translates each question and each answer option of the 980 items into the six languages; the answer instruction is a fixed sentence written in each language.
Before translation, numerals, math symbols, unit tokens, and option labels (A--D) are masked with placeholders and restored afterwards. Outputs are not post-edited, so each language condition is a \emph{machine-translated} question.

\paragraph{What the agreement check measures.}
For 50 random items per language (seed 42) we translated the English question independently with Gemini~2.0 Flash and compared the two outputs.
This is a forward-translation agreement check between two MT systems on the question text only (options and instructions were not compared); no human annotators were involved in this check.
The agreement score (0.79--0.84, Table~\ref{tab:languages}) is a set-based character Jaccard index: it compares only the sets of distinct characters in the two strings, ignoring order and repetition.
Unrelated question pairs in the same language already score 0.45--0.52 on average, so the score is only a coarse screen (Appendix~\ref{app:translation}).
Sentence-level chrF~\cite{popovic-2015-chrf} is more informative (59--68 for matched pairs vs.\ 12--16 for unrelated ones; Table~\ref{tab:languages}), but agreement between two MT systems still cannot establish that terminology, quantities, and logical structure survive translation.

\paragraph{Fidelity audit.}
We therefore audited all $980\times6$ translations automatically (Appendix~\ref{app:translation}).
Per language, 23.6--24.5\% of items carry a flag for numeric, notation, or structural changes (15.0--16.0\% in the question text, 11.3--11.7\% in the options), including a formatting change to the ``Yes'' option of all 77 yes/no items.
Flags are concentrated in MMMU and nearly identical across languages: 194 items are flagged in all six languages.
An LLM-assisted triage of the 300 verification samples, not yet confirmed by native speakers, finds a major meaning error or omission in 19\% of them (95\% CI 15--24\%).
The reported English-to-Indian drops therefore combine a language effect with machine-translation noise, so they are best read as an upper bound on the pure language effect.
We release the per-item flags, a clean-subset list, and a native-speaker validation protocol.

\subsection{Models}
Table~\ref{tab:models} lists the eight VLMs: GPT-4o~\cite{openai2024gpt4o} (proprietary) and seven open-weight models, Qwen3-VL-30B~\cite{bai2025qwen3vl}, Gemma~3-27B~\cite{google2025gemma3}, Llama-4-Maverick~\cite{meta2025llama4}, Qwen2.5-VL-7B and -32B~\cite{bai2025qwen25vl}, InternVL2.5-8B~\cite{chen2024internvl25}, and Aya-Vision-8B~\cite{dash2025ayavision}.
Qwen2.5-VL-7B, InternVL2.5-8B, and Aya-Vision-8B are served with vLLM on an AMD Instinct MI300X GPU (192\,GB) in bfloat16; Qwen3-VL-30B, Gemma~3-27B, Llama-4-Maverick (FP8), and Qwen2.5-VL-32B are queried through DeepInfra and GPT-4o through the OpenAI API (snapshot March 2026); images are capped at $512{\times}512$ pixels.
On Qwen2.5-VL-7B only, we additionally run a \textbf{no-image} condition (image removed, text-only) and a \textbf{CoT} condition in which the answer instruction is replaced by its translated ``Think step by step, then answer \ldots'' variant; we chose it as a representative open-weight model that we serve locally, which gives full control over prompting and decoding. Ablation results are therefore specific to this model and prompt.

\subsection{Evaluation Protocol}
\label{sec:eval}
Each model receives the translated question, the translated options labelled with Latin letters A, B, C, \dots, the original image, and an instruction in the prompt language asking for the option letter (A--D) or a number, at temperature 0.
For MCQs (762 of 980) we take the first standalone Latin letter A--D, falling back to the last capital A--D anywhere in the response; option letters written in Indian scripts, yes/no words, and option text are not recognized.
For free-form numerical items (218 of 980), Devanagari, Bengali, Tamil, Telugu, and Kannada digits are converted to Arabic digits, the last number is taken, and a $\pm$5\% tolerance is applied.
Responses with no extractable letter or number are scored as incorrect; failure rates are substantial for some models (33.9\% of all Aya-Vision-8B responses, 25.0\% for InternVL2.5-8B), are higher in Dravidian than in Indo-Aryan conditions for six of eight models, and are analysed in Section~\ref{sec:extraction-sens} and Appendix~\ref{app:extraction}.
We compute 95\% percentile-bootstrap confidence intervals (2{,}000 resamples) and paired McNemar tests between English and each Indian language ($p{<}0.001$ for all 48 model--language pairs); aggregated drops and family gaps use unrounded accuracies.
Refusals are flagged by matching English phrases (``cannot'', ``I don't''), so refusals written in Indian languages are not detected.

\subsection{What the Design Does and Does Not Isolate}
\label{sec:scope}
Holding each image fixed removes visual content as an explanation for differences on the same question, and every model sees identical questions in every language.
The manipulated variable, however, is the whole \emph{text condition}, not language in the abstract.
A lower score in a language bundles at least (i) the model's reasoning when prompted in that language; (ii) machine-translation fidelity (Section~\ref{sec:translation}); (iii) tokenizer coverage and sequence length for the script; (iv) morphological and typological distance from English; (v) the model's pretraining and post-training exposure to the language, which is rarely documented; (vi) whether an answer can be extracted from the response (Section~\ref{sec:eval}); and (vii) interaction with English text left in the unmodified images (43.2\% of items).
Language family and script covary with all of these, and our design has three languages per family and a single same-script pair.
We therefore report per-language drops and family differences as measured regularities, and we label explanations for them (training-data volume, reasoning transfer, post-training alignment) as hypotheses for future controlled tests.

\section{Results}
\label{sec:results}

\subsection{Overall Accuracy}
Table~\ref{tab:main} reports per-language accuracy.
GPT-4o achieves the highest Indian-language mean accuracy (43.2\%), while Gemma~3-27B, despite lower English accuracy, drops only 9.8$\pp$ -- the best cross-lingual robustness among all models tested.
At the other extreme, Qwen3-VL-30B posts the highest English accuracy (64.9\%) but loses 24.9$\pp$ on Indian languages, showing that strong English performance does not automatically transfer cross-lingually.
Among the 7--8B open models, drops range from 20.4$\pp$ (InternVL2.5-8B) to 21.9$\pp$ (Aya-Vision-8B).
Refusal rates, detected by matching English refusal phrases (``cannot'', ``I don't''), stay under 3\% for all eight models in every language (maximum 2.45\%); because this detector cannot recognize refusals written in Indian languages, it does not by itself show that the losses are reasoning failures, and answer-extraction failures (Section~\ref{sec:extraction-sens}) are a larger confound.

\begin{table*}[t]
\centering
\small
\setlength{\tabcolsep}{4pt}
\caption{Accuracy (\%) per model and language on the 980 items (image fixed, question text translated). \textbf{Bold}: best model per column (for Drop: smallest). Ind.\ mean = mean of the six Indian languages; Drop = EN $-$ Ind.\ mean. Takeaway: every model is less accurate in every Indian language than in English (McNemar $p{<}0.001$ for all 48 model--language pairs); GPT-4o is best on the Indian-language mean, Qwen3-VL-30B is best in English but has the largest drop, and Gemma~3-27B has the smallest drop. 95\% bootstrap CIs (half-widths 1.9--3.3\,pp) are in Table~\ref{tab:ci}.}
\label{tab:main}
\begin{tabular}{@{}l c ccc ccc cc@{}}
\toprule
 & & \multicolumn{3}{c}{\textit{Indo-Aryan}} & \multicolumn{3}{c}{\textit{Dravidian}} & & \\
\cmidrule(lr){3-5}\cmidrule(lr){6-8}
\textbf{Model} & \textbf{EN} & \textbf{hi} & \textbf{bn} & \textbf{mr} & \textbf{ta} & \textbf{te} & \textbf{kn} & \textbf{Ind.\ mean} & \textbf{Drop} \\
\midrule
GPT-4o & 54.5 & \textbf{44.6} & \textbf{44.5} & \textbf{43.2} & 40.5 & \textbf{43.0} & \textbf{43.7} & \textbf{43.2} & 11.3 \\
Gemma~3-27B & 51.8 & 42.0 & 41.9 & 42.2 & \textbf{41.9} & 42.6 & 41.3 & 42.0 & \textbf{9.8} \\
Llama-4-Maverick & 55.4 & 43.4 & 42.7 & 41.9 & 34.9 & 38.6 & 41.6 & 40.5 & 14.9 \\
Qwen3-VL-30B & \textbf{64.9} & 44.1 & 43.3 & 36.1 & 41.1 & 36.5 & 38.7 & 40.0 & 24.9 \\
Qwen2.5-VL-32B & 50.3 & 34.1 & 39.1 & 35.6 & 34.2 & 33.2 & 32.6 & 34.8 & 15.5 \\
Qwen2.5-VL-7B & 51.3 & 34.2 & 32.7 & 30.0 & 28.7 & 26.9 & 30.6 & 30.5 & 20.8 \\
InternVL2.5-8B & 45.6 & 29.8 & 24.9 & 27.9 & 22.3 & 22.9 & 23.6 & 25.2 & 20.4 \\
Aya-Vision-8B & 42.0 & 33.6 & 21.7 & 25.1 & 14.9 & 11.6 & 14.2 & 20.2 & 21.9 \\
\midrule
\multicolumn{10}{@{}l}{\textit{Qwen2.5-VL-7B prompt ablations (single model; not included in bolding)}} \\
No-image & 35.8 & 26.0 & 24.5 & 25.1 & 19.0 & 20.7 & 23.7 & 23.2 & 12.7 \\
Chain-of-thought & 49.5 & 25.0 & 18.3 & 33.3 & 29.7 & 22.8 & 19.2 & 24.7 & 24.8 \\
\bottomrule
\end{tabular}
\end{table*}

\subsection{Sensitivity to Answer Extraction}
\label{sec:extraction-sens}
Because unparsed answers count as wrong, a measured drop mixes reasoning with answer-format effects, and extraction failures are language-dependent: for Aya-Vision-8B they rise from 9.1\% in English to 21.9\% (Indo-Aryan) and 54.3\% (Dravidian), and for InternVL2.5-8B from 16.2\% to 23.5\% and 29.6\%.
We therefore recompute accuracies under two further assumptions (Table~\ref{tab:family}, Appendix~\ref{app:extraction}): S2 scores only validly extracted responses, and S3 credits failures at chance ($p{=}0.284$).
The Dravidian drop still exceeds the Indo-Aryan drop for seven of eight models under S2 (Qwen3-VL-30B reverses to $-0.5\pp$) and all eight under S3, but the gap shrinks: Aya-Vision-8B's 13.2\pp{} falls to 4.1--6.2\pp, InternVL2.5-8B's 4.6\pp{} to 2.9--3.3\pp, and Qwen2.5-VL-7B's 3.5\pp{} to 2.1--2.5\pp.
We therefore read the family effect as consistent in direction but modest, and a hidden accuracy of about 0.4 among failed responses would erase the gaps of Aya-Vision-8B and Qwen3-VL-30B; separating answer format from reasoning requires a manual audit of failed responses, for which we release a protocol and script (Appendix~\ref{app:extraction}).

\subsection{Language Family Effect}
\begin{figure}[!ht]
\centering
\includegraphics[width=\columnwidth]{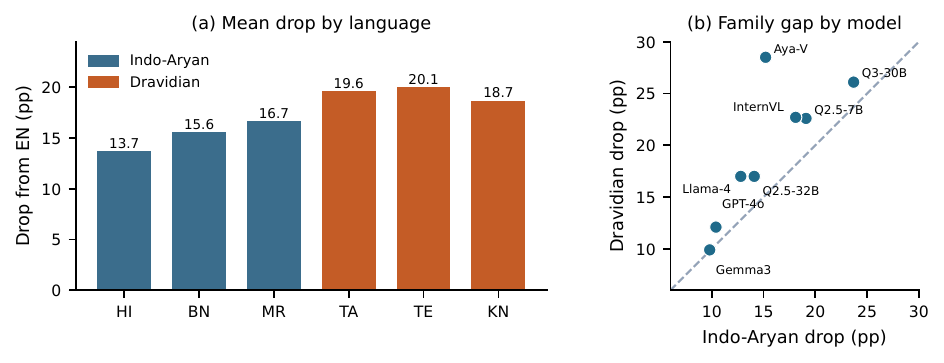}
\caption{Dravidian languages lose more than Indo-Aryan languages for every model, by very different amounts (cf.\ Table~\ref{tab:family}). (a) Mean drop from English per language over the eight models: 18.7--20.1$\pp$ for the Dravidian languages vs.\ 13.8--16.7$\pp$ for the Indo-Aryan ones. (b) Indo-Aryan vs.\ Dravidian drop per model; all points lie on or above the diagonal, from Gemma~3-27B (+0.1$\pp$) to Aya-Vision-8B (+13.2$\pp$).}
\label{fig:panel}
\end{figure}

Table~\ref{tab:family} and Figure~\ref{fig:panel} disaggregate the drop by language family for all eight models.
Under strict scoring the Dravidian drop exceeds the Indo-Aryan drop for every model, but the gap varies widely: 0.1\pp{} for Gemma~3-27B, 1.7--4.6\pp{} for six models, and 13.2\pp{} for Aya-Vision-8B; per model it is significant ($p<0.05$, two-sample $t$-test over the 3+3 per-language accuracies) only for Aya-Vision-8B and InternVL2.5-8B.
Aya-Vision-8B's declared languages include Hindi but none of our other five~\cite{ayavisioncard}, so its large gap is compatible with limited language coverage, and it is also the model whose Dravidian responses most often fail extraction (54.3\% vs.\ 21.9\%; Section~\ref{sec:extraction-sens}).
Figure~\ref{fig:panel}(b) plots Indo-Aryan vs.\ Dravidian drops per model; under strict scoring all points lie on or above the diagonal.

\begin{table}[t]
\centering
\small
\setlength{\tabcolsep}{3.5pt}
\caption{Mean accuracy drop from English (pp) by language family for all eight models, sorted by the strict gap (Dravidian $-$ Indo-Aryan drop, from unrounded accuracies). S1 = strict scoring (extraction failures wrong; used throughout); S2 = accuracy among validly extracted responses only; S3 = failures credited at chance ($p{=}0.284$); see Section~\ref{sec:extraction-sens}. $^\dagger$\,$p<0.05$, two-sample $t$-test over the 3+3 per-language accuracies (exploratory). Takeaway: the Dravidian drop exceeds the Indo-Aryan drop for 8/8 models under S1 and S3 and 7/8 under S2, but outside S1 no gap exceeds 4.5\,pp; most of Aya-Vision-8B's 13.2\,pp depends on how its extraction failures are scored.}
\label{tab:family}
\begin{tabular}{@{}lcc rrr@{}}
\toprule
 & & & \multicolumn{3}{c}{\textbf{Gap}} \\
\cmidrule(l){4-6}
\textbf{Model} & \textbf{Indo-A.} & \textbf{Drav.} & \textbf{S1} & \textbf{S2} & \textbf{S3} \\
\midrule
Aya-Vision-8B & 15.2 & 28.5 & +13.2$^\dagger$ & +4.5 & +4.1 \\
InternVL2.5-8B & 18.1 & 22.7 & +4.6$^\dagger$ & +3.3 & +2.9 \\
Llama-4-Maverick & 12.8 & 17.0 & +4.3 & +2.8 & +3.2 \\
Qwen2.5-VL-7B & 19.0 & 22.6 & +3.5 & +2.1 & +2.1 \\
Qwen2.5-VL-32B & 14.0 & 17.0 & +3.0 & +4.4 & +3.5 \\
Qwen3-VL-30B & 23.7 & 26.1 & +2.4 & $-$0.5 & +0.7 \\
GPT-4o & 10.4 & 12.1 & +1.7 & +1.5 & +1.5 \\
Gemma~3-27B & 9.8 & 9.9 & +0.1 & +0.3 & +0.2 \\
\bottomrule
\end{tabular}
\end{table}

\paragraph{Script versus language.}
Hindi and Marathi share Devanagari, yet Marathi is less accurate than Hindi for six of eight models, by 3.0\pp{} on average (3.1\pp{} when only validly extracted responses are scored; median 1.7\pp); the difference is concentrated in Aya-Vision-8B (8.5\pp) and Qwen3-VL-30B (8.0\pp).
Because script is held fixed, this points to language-specific factors such as training-data volume, tokenizer coverage, or translation quality; with a single same-script pair and no per-language pretraining statistics, our design cannot test these hypotheses.
Translation-quality proxies do not follow the family pattern: automated defect rates are nearly flat across languages (23.6--24.5\%), and chrF ranks Tamil (65.1) above Bengali (62.9) although Tamil accuracy is at or below Bengali accuracy for all eight models. Translation noise alone therefore does not explain the Dravidian--Indo-Aryan gap, but it is not ruled out either: the LLM-assisted triage marks more Kannada (30\%) than Hindi (12\%) samples as major errors or omissions, with overlapping intervals (Appendix~\ref{app:translation}).

\subsection{Per-Dataset Patterns}
Table~\ref{tab:perdataset} breaks drops down by source benchmark.
MathVista produces the largest drop for the three strongest models (29.3$\pp$ for Qwen3-VL-30B) and ScienceQA for the other five (up to 38.4$\pp$ for Aya-Vision-8B and 36.8$\pp$ for InternVL2.5-8B); one possible reason for the former is that math word problems depend on precise comprehension of quantities and relations stated in the text.
Gemma~3-27B is slightly \emph{more} accurate on ScienceQA in Indian languages than in English ($-$2.7$\pp$ drop); its training recipe is not public, and we do not attribute this to any specific cause.
The large ScienceQA drops of Aya-Vision-8B and InternVL2.5-8B could reflect reliance on English science vocabulary, but they coincide with the two highest Indian-language extraction-failure rates, so we do not interpret them further; ScienceQA is letter-keyed multiple choice, and its family gap is positive for all eight models (+1.7 to +24.1\pp; Appendix~\ref{app:extraction}).

\begin{table}[t]
\centering
\small
\caption{Accuracy drop (pp) from English to the mean of the six Indian languages, by source benchmark, for all eight models ($n$ = 400 / 400 / 180). \textbf{Bold}: smallest drop per column; negative values mean Indian-language accuracy above English. Takeaway: ScienceQA and MMMU drops vary widely across models (Gemma~3-27B has none on ScienceQA). MathVista drops should be read with the scoring caveat in the Limitations section.}
\label{tab:perdataset}
\begin{tabular}{@{}lccc@{}}
\toprule
\textbf{Model} & \textbf{MathVista} & \textbf{ScienceQA} & \textbf{MMMU} \\
\midrule
GPT-4o & 17.2 & 7.0 & 7.3 \\
Gemma~3-27B & 24.3 & \textbf{$-$2.7} & \textbf{5.4} \\
Llama-4-Maverick & 15.2 & 15.7 & 12.3 \\
Qwen3-VL-30B & 29.3 & 21.5 & 22.8 \\
Qwen2.5-VL-32B & 10.2 & 23.2 & 10.3 \\
Qwen2.5-VL-7B & 20.5 & 24.8 & 12.9 \\
InternVL2.5-8B & \textbf{8.1} & 36.8 & 11.4 \\
Aya-Vision-8B & 9.3 & 38.4 & 13.1 \\
\bottomrule
\end{tabular}
\end{table}

\subsection{Cross-Lingual Consistency}
For each question with a valid extracted answer in at least four of the seven languages, we take the most frequent (modal) answer and compute the share of valid answers equal to it; a model's \emph{consistency} is the mean of this share over such questions (100\% means the same answer in every language, whether correct or not).
Consistency ranges from 67.1\% (Aya-Vision-8B) to 80.8\% (Llama-4-Maverick) (Figure~\ref{fig:consistency}).
If every valid answer were an independent uniform guess among the (at most four extractable) options, our MCQs would give 55.2\% (seven valid answers) to 60.8\% (four valid answers), so all models are above chance, Aya-Vision-8B only modestly.
Consistency and accuracy drop are only loosely related: Qwen3-VL-30B is fairly consistent (78.3\%) yet has the largest drop (24.9\pp).

\subsection{Prompt Ablations (Qwen2.5-VL-7B only)}
\label{sec:ablations}
\begin{figure*}[!t]
\centering
\begin{subfigure}[t]{0.48\textwidth}
\centering
\includegraphics[width=\linewidth]{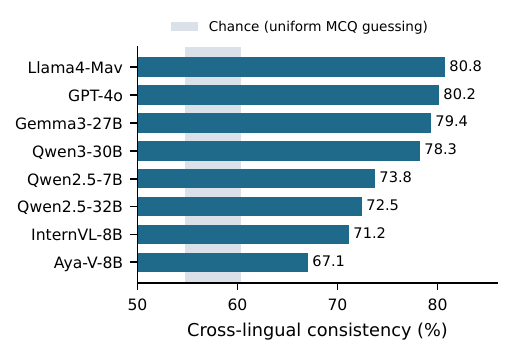}
\caption{Cross-lingual consistency (mean share of a question's valid answers equal to its modal answer; $\geq$4 of 7 valid). All models (67.1--80.8\%) exceed the 55.2--60.8\% expected from uniform guessing (shaded).}
\label{fig:consistency}
\end{subfigure}\hfill
\begin{subfigure}[t]{0.48\textwidth}
\centering
\includegraphics[width=\linewidth]{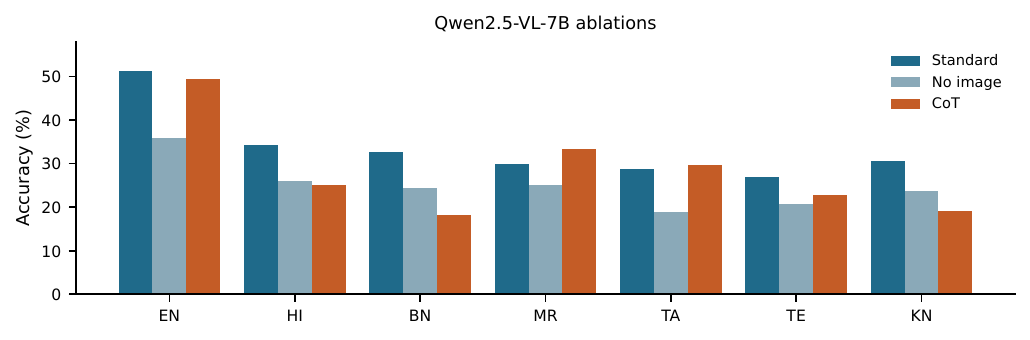}
\caption{Qwen2.5-VL-7B only: removing the image hurts English most ($-$15.5$\pp$); CoT hurts Bengali, Kannada, and Hindi ($-$9.2 to $-$14.4$\pp$) but barely changes English.}
\label{fig:ablation}
\end{subfigure}
\caption{(a) Models give the same answer across languages more often than chance, but consistency does not track accuracy (Qwen3-VL-30B is 78.3\% consistent yet has the largest drop). (b) Ablation effects hold for one model and one prompt, and the CoT losses coincide with 13--17$\pp$ more extraction failures (Section~\ref{sec:ablations}).}
\end{figure*}

Figure~\ref{fig:ablation} compares standard, no-image, and CoT prompts on Qwen2.5-VL-7B, the only model we ablate; all statements below are specific to this model and prompt.
Removing the image reduces English accuracy by 15.5\pp{} (51.3\%$\to$35.8\%) but Indian-language accuracy by only 4.9--9.7\pp. This asymmetry is partly a response-format effect: without an image, 26.2\% of English responses contain an English refusal phrase and English extraction failures rise from 6.6\% to 24.8\%; scoring only validly extracted responses, the English cost is 7.3\pp, within the Indian range (3.8--11.2\pp).
Replacing the answer instruction with a translated ``Think step by step, then answer \ldots'' instruction changes English by $-$1.8\pp{} but lowers Bengali ($-$14.4\pp), Kannada ($-$11.4\pp), and Hindi ($-$9.2\pp), while Marathi gains 3.3\pp. These losses track extraction failures, which rise by 13.4, 17.0, and 13.0\pp{} in the same three languages and fall for Marathi and Tamil, so part of the effect may be answer-format drift in longer outputs rather than degraded reasoning; CoT still lowers accuracy in five of six Indian languages when only valid extractions are scored.
The direction resembles the text-only finding of \citet{shi2023mgsm}, but we did not test English CoT with Indian-language questions or any other model.

\subsection{Effect of Scale}
Comparing Qwen2.5-VL-7B with Qwen2.5-VL-32B (same family) shows the larger model gains 4.3$\pp$ on Indian languages while losing only 1.0$\pp$ on English; the English-to-Indian drop shrinks from 20.8$\pp$ to 15.5$\pp$.
This is a meaningful reduction but far from closing the gap.
Part of the comparison is confounded by answer extraction: Qwen2.5-VL-32B fails extraction on 21.3\% of English responses versus 6.6\% for the 7B model, which depresses its English score (Section~\ref{sec:mcq}).
With one model family and two sizes, we cannot tell whether scale mainly broadens language coverage or also improves cross-lingual reasoning transfer.

\section{Analysis}
\label{sec:analysis}
We complement accuracy tables with three response-level surface diagnostics (Figures~\ref{fig:leak}--\ref{fig:length}): English-word share, script mismatch, and response length. They describe outputs that accuracy alone does not reveal, but they do not by themselves identify a model's internal reasoning language.

\subsection{English Token Leak and Script Confusion}
\begin{figure*}[!t]
\centering
\begin{subfigure}[t]{0.48\textwidth}
\centering
\includegraphics[width=\linewidth]{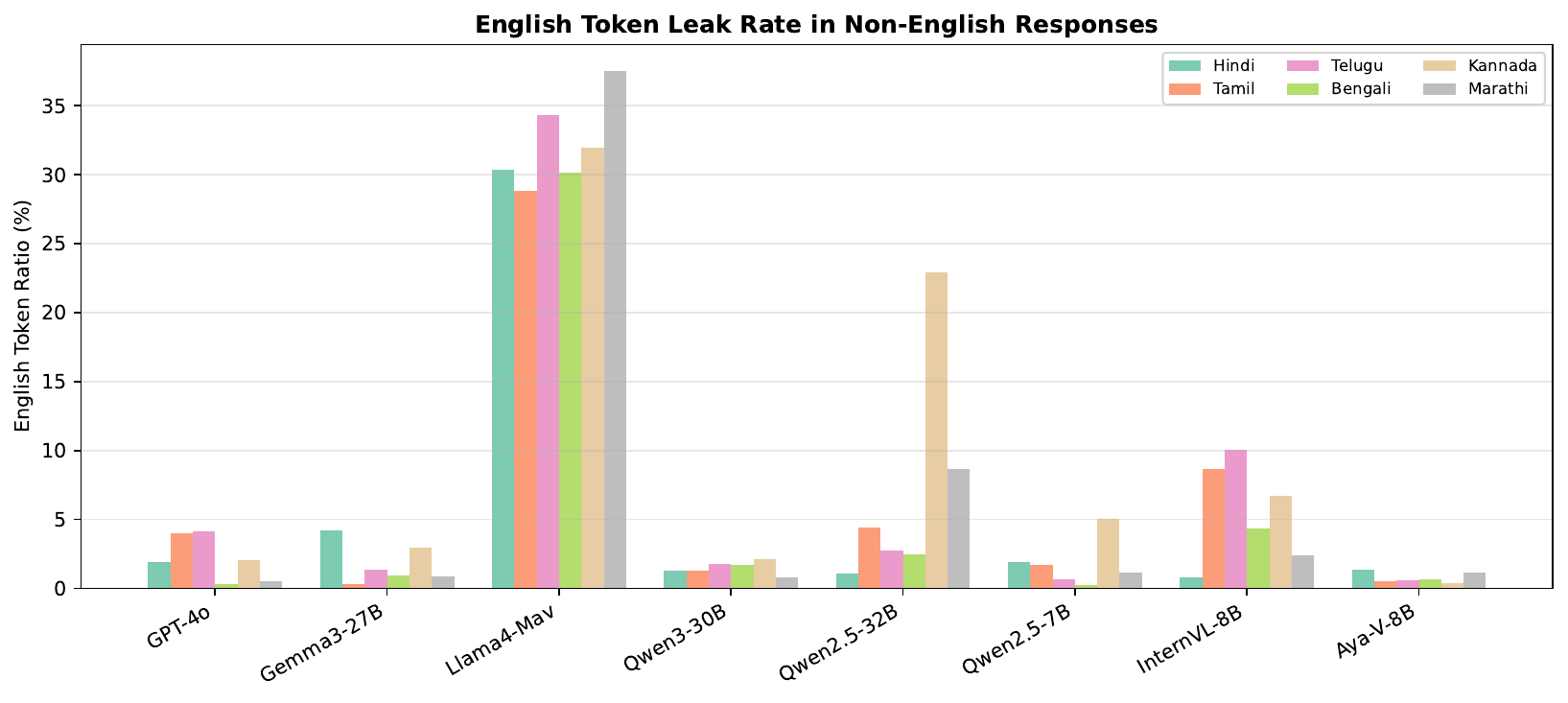}
\caption{Share of English words ($\geq$3 letters) in responses to Indian-language prompts: Llama-4-Maverick is the outlier (${\sim}$32\% on average), Aya-Vision-8B the lowest (0.8\%).}
\label{fig:leak}
\end{subfigure}\hfill
\begin{subfigure}[t]{0.48\textwidth}
\centering
\includegraphics[width=\linewidth]{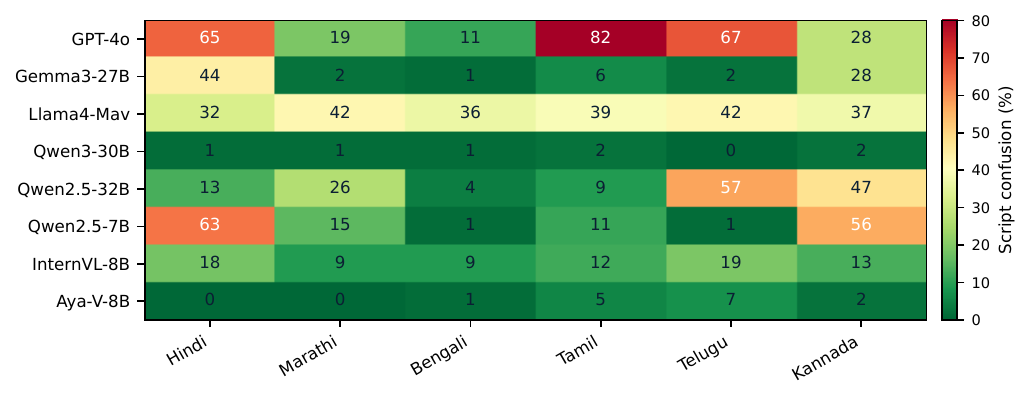}
\caption{Share of responses longer than 15 characters with no target-script character: 45\% on average for GPT-4o, near zero for Qwen3-VL-30B and Aya-Vision-8B.}
\label{fig:confusion}
\end{subfigure}
\caption{Models with similar accuracy differ sharply in output language. Llama-4-Maverick mixes in English yet is the most consistent model; GPT-4o's high script-mismatch rate mostly reflects Latin-script letter answers (e.g., ``The answer is B.''), which our instruction allows. These are surface measures of the output, not of internal reasoning.}
\end{figure*}

As a surface proxy for English-mediated answering, we measure the fraction of $\geq$3-letter English words in responses to Indian-language prompts (Figure~\ref{fig:leak}).
Llama-4-Maverick stands out with $\sim$32\% English tokens on average across Indian-language responses, yet still achieves the highest cross-lingual consistency (80.8\%).
Aya-Vision-8B shows only 0.8\% English tokens but the lowest consistency (67.1\%).
We also flag responses longer than 15 characters containing no characters from the expected Unicode script range (Figure~\ref{fig:confusion}).
Qwen3-VL-30B (1.2\%) and Aya-Vision-8B (2.5\%) stay on-script; GPT-4o's 45\% rate mostly reflects Latin-script answers (e.g., ``The answer is B.''), which our letter-answer instruction permits, so this measure flags script choice rather than comprehension failure.

\subsection{Response-Length Patterns}
\begin{figure}[!ht]
\centering
\includegraphics[width=\columnwidth]{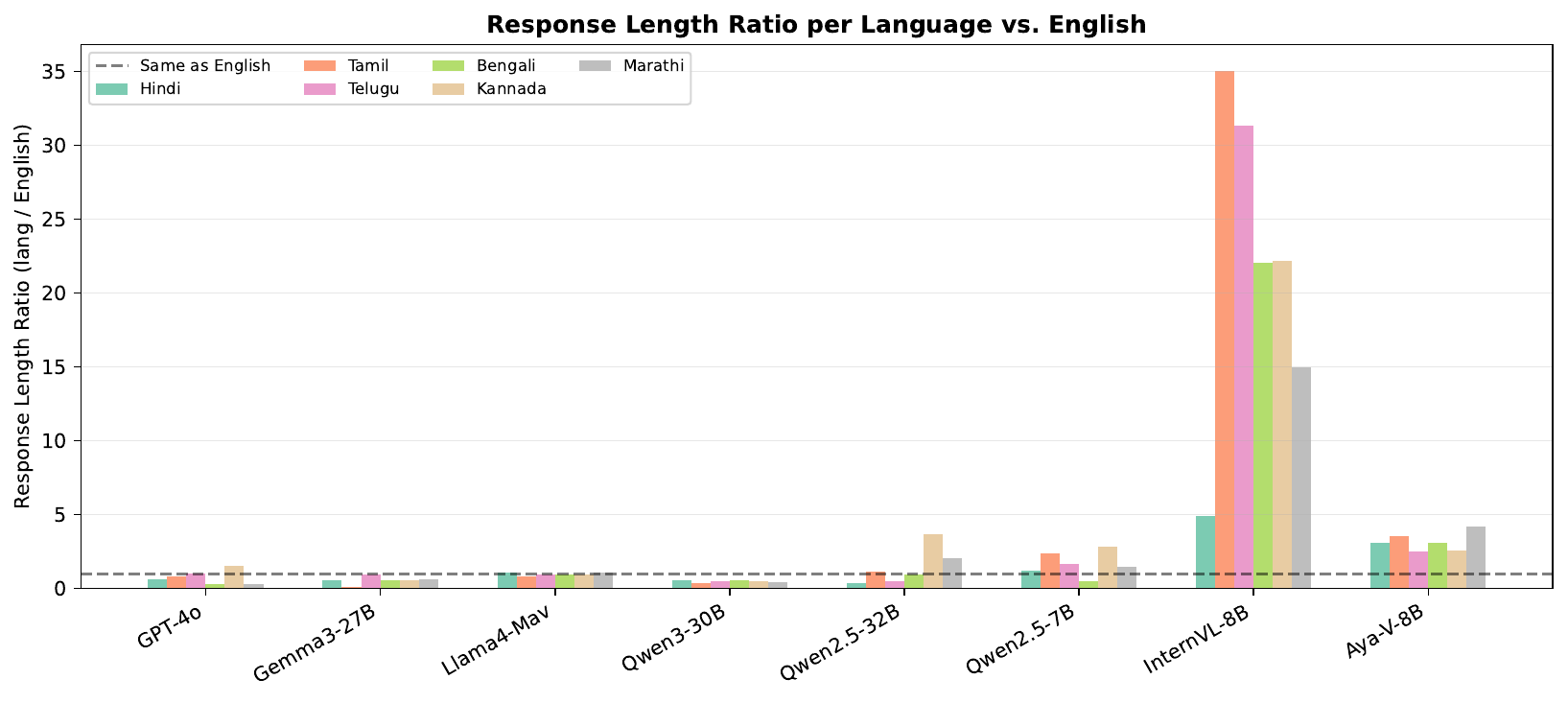}
\caption{Mean response length in each Indian language relative to English (dashed line = equal length). GPT-4o, Gemma~3-27B, Llama-4-Maverick, and Qwen3-VL-30B stay at or below parity; Aya-Vision-8B writes 2.5--4$\times$ longer Indian-language responses (310--514 vs.\ 123 characters), and InternVL2.5-8B is off the scale because its English answers are single letters (6 characters on average) while its Indian-language answers are full sentences.}
\label{fig:length}
\end{figure}

Response length reveals strategy differences across models (Figure~\ref{fig:length}).
Llama-4-Maverick writes uniformly verbose output ($\sim$840--1{,}070 characters) regardless of language, always producing full reasoning.
Aya-Vision-8B produces \emph{longer} responses in Indian languages (310--514 characters) than in English (123 characters); we do not know why, and lower confidence is only one possibility.
InternVL2.5-8B shows the reverse: 6-character English responses (single-letter MCQ answers) but full-sentence Indian-language output, which aligns with its mixed script-confusion pattern in Figure~\ref{fig:confusion} despite moderate mean rates.

\subsection{MCQ versus Free-Form Questions}
\label{sec:mcq}
Free-form numerical items suffer disproportionately.
Qwen2.5-VL-32B reaches 63.9\% on MCQs but only 2.8\% on free-form numerical items in English, generating extended reasoning without a clean final number and triggering extraction failures -- a 61$\pp$ gap, the largest we observe, that highlights how strongly reported accuracy depends on answer-extraction quality.
InternVL2.5-8B shows a similar, smaller pattern (52.5\% MCQ vs.\ 21.6\% free-form).
This gap reflects answer formatting rather than reasoning, and it depresses Qwen2.5-VL-32B's English score in the scale comparison.

\section{Discussion}
\label{sec:discussion}

\paragraph{Implications for evaluation.}
Our results do not show that any model is suitable for tutoring in any language: English accuracy is only 42--65\%.
What they show is that English-only evaluation understates the difficulty of the same questions in Indian languages by 9.8--24.9$\pp$, by more for Dravidian languages in every model, and by amounts that differ enough across models to reorder them: Qwen3-VL-30B is first in English but fourth on the Indian-language mean, whereas Gemma~3-27B loses only 9.8$\pp$.
NEP~2020 envisages home-language instruction until at least Grade~5 and bilingual science and mathematics materials~\cite{nep2020}; any future system for such settings would first need per-language evaluation of this kind, with translation and answer-extraction controls.

\paragraph{Massively multilingual training does not guarantee coverage.}
Aya-Vision-8B is trained for 23 languages, but among our six only Hindi is one of them~\cite{ayavisioncard}.
Consistent with this, it is most accurate in Hindi (33.6\%) and least accurate in the three Dravidian languages (11.6--14.9\%), while still answering in the target script (0.8\% English words) and rarely refusing.
Its large Dravidian penalty is therefore compatible with limited exposure to those languages, not only with a failure of reasoning transfer, and more than half of its Dravidian responses fail answer extraction.
Separating these explanations would require per-language training-data statistics and models whose declared coverage includes the Dravidian languages.

\paragraph{Gemma~3-27B as an outlier.}
Gemma~3-27B is the only model with a near-zero Dravidian--Indo-Aryan gap (0.1$\pp$) and is slightly more accurate on ScienceQA in Indian languages than in English.
Its training recipe is not public, so we cannot attribute this to any specific choice; multilingual alignment during post-training is one hypothesis that could be tested with models whose training data are documented.

\paragraph{Chain-of-thought: a single-model caution.}
For Qwen2.5-VL-7B with one translated ``think step by step'' instruction, CoT lowered accuracy in four of six Indian languages (Figure~\ref{fig:ablation}) while leaving English nearly unchanged; Marathi (+3.3$\pp$) and Tamil (+1.0$\pp$) improved.
This design cannot tell whether the losses come from reasoning in the target language or from longer outputs that our extractor fails to parse: the three languages with the largest losses also show 13--17$\pp$ more extraction failures, whereas Marathi and Tamil show fewer.
Whether CoT hurts other models, whether English CoT with Indian-language questions behaves differently, and whether constrained answer formats remove the effect are open questions, so we do not generalize beyond this model and prompt.

\paragraph{The hidden-English pattern.}
Llama-4-Maverick combines about 32\% English words in its Indian-language responses (Figure~\ref{fig:leak}) with the highest cross-lingual consistency (80.8\%; Figure~\ref{fig:consistency}).
One hypothesis is that it reasons largely in English and translates only at the surface; our surface statistics cannot confirm this, which would require probing intermediate representations or constraining the output language.
If true, it would matter more for generative uses such as explanations or feedback, which need fluent target-language text, than for letter-answer MCQs.

\section{Conclusion}
\label{sec:conclusion}

Across 980 questions, seven languages, eight VLMs, and 68{,}600 inference records, we measure a 9.8--24.9$\pp$ accuracy penalty when the same diagram-grounded STEM questions are posed in Indian languages rather than English, a larger penalty for Dravidian than for Indo-Aryan languages in every model, and, for Qwen2.5-VL-7B with our prompt, CoT losses in four of six Indian languages that coincide with more extraction failures.
Neither a 23-language model whose coverage includes only Hindi among our languages nor scaling Qwen2.5-VL from 7B to 32B closes the gap.
Part of each measured gap may come from machine translation, answer extraction, and English text in the unmodified images, which we quantify but cannot fully remove.
We recommend that multilingual VLM evaluations report per-language accuracy together with these three controls, rather than English-only or pooled scores.
Translations, images, and code are publicly available (footnote~1) to support reproduction and extension to further languages.

\section*{Limitations}
\textbf{Translation.} All non-English questions are machine translations. Our automatic agreement check compares IndicTrans2 with an independent Gemini~2.0 Flash translation on 50 questions per language using character-set Jaccard similarity (0.79--0.84), which cannot detect many terminology, unit, negation, or spatial-relation errors; an automated audit flags numeric, notation, or option changes in 23.6--24.5\% of items per language, and an LLM-assisted triage finds major errors or omissions in 19\% of 300 samples. \emph{Next steps:} native-speaker review of a stratified sample per language with re-scoring on the verified subset, and a comparison of translators to bound translation-induced variance.
\textbf{Images.} Images are unmodified, and 43.2\% contain English text, so the Indian-language conditions are mixed-language. This could shrink the measured gap, because models can fall back on English text in the image, or widen it, because models must match translated terms to English labels (in 29.7\% of items a content word of the question also appears as an English label in the image). \emph{Next step:} rendering translated text back into the figures, by typesetting labels for synthetic charts and diagrams or by scene-text inpainting for photographs, to test whether fully localized items widen or narrow the gap.
\textbf{Answer extraction.} Unparsed responses count as wrong; failures reach 33.9\% (Aya-Vision-8B) and 25.0\% (InternVL2.5-8B) and are higher for Dravidian languages (54\% vs.\ 22\% for Aya-Vision-8B), so part of the family gap may reflect answer format rather than reasoning; the family gap persists in direction but shrinks to at most 4.5\pp{} when failures are excluded and 4.1\pp{} when they are credited at chance. Refusals are detected with English phrases only. \emph{Next steps:} a manually validated extraction sample and a permissive, script-aware extractor applied to the per-response outputs.
\textbf{Causal scope.} Language family is confounded with tokenizer coverage, typological distance from English, morphology, translation quality, and pretraining exposure (Section~\ref{sec:scope}), so family and script findings are descriptive; the ablations use one model and one prompt. \emph{Next steps:} controls for tokenization and documented per-language training data, and ablations on several models with both English and target-language CoT.
\textbf{Coverage.} We use 980 questions (180 from MMMU) selected by filters and dataset order rather than stratified sampling, six of the 22 scheduled languages, and English-source benchmarks that may carry curricular or cultural biases; low-resource languages such as Odia and Assamese and the 99 non-scheduled languages recorded in the 2011 Census remain untested.
\textbf{Snapshot.} Hosted models may change after our March 2026 snapshot; results are a snapshot rather than a permanent ranking.

\paragraph{Scoring caveats.}
Two answer-format properties of the scoring, independent of the models, bound the interpretation of some drops.
First, 84 items (69 MathVista yes/no questions posed with options, 10 MCQs keyed to option E or later, and 5 free-form items with non-numeric keys) cannot be credited in any language.
Second, 114 MathVista MCQs are keyed by option text, and translation alters the option strings (e.g., ``6cm''$\to$``6 cm''), so our API-pipeline scorer credits these items in English but only 24--25 of them in each Indian language; this can inflate the English$\to$Indian drop of API-evaluated models by up to 9.2\pp{} (for Qwen2.5-VL-32B, at least 5.7 of its 15.5\pp), whereas our local-inference scorer credits none of the 114 in any language, which lowers but does not skew those models' accuracies.
Because the number of creditable items differs by at most one across the six Indian languages, family-gap and Hindi--Marathi comparisons are unaffected; English-to-Indian drops of API-served models, and MathVista drops in particular, should be read as upper bounds.

\paragraph{Input format.}
For 162 of the 167 MMMU multiple-choice items, all answer choices are presented as a single option line (e.g., \texttt{A.~['Yes.', 'No.', 'Not sure.']}); eleven multi-image MMMU items keep only their first image; the ScienceQA context field is omitted (it is non-empty for 376 of 400 items); two MathVista source items (mathvista\_204, mathvista\_243) are in Persian rather than English; and unit-preserving masking alters the ``Yes'' option string of all 77 yes/no items (e.g., Hindi ``ye~s''). These affect every language condition, but they make items differ from the original benchmarks, and the translation-side changes add to the English-to-Indian drop.

\section*{Ethical Considerations}
\textbf{Intended use.} This is an evaluation study. The best English accuracy we observe is 64.9\%, far below what student-facing use would require, and nothing here endorses deploying these models in classrooms in any language.
\textbf{Data and licenses.} We translate and redistribute questions and images derived from MathVista (CC BY-SA 4.0), ScienceQA (CC BY-NC-SA 4.0), and MMMU (Apache 2.0); derived files follow the most restrictive applicable terms (non-commercial and share-alike for ScienceQA-derived items). The translations are machine-generated and intended for evaluation research.
\textbf{Human subjects.} No human annotators or participants were involved.

\section*{Acknowledgments}
The author thanks AMD for providing access to the AMD Instinct MI300X GPU used for open-model inference in this work, and the MRL reviewers for comments that strengthened the translation, extraction, and framing analyses. An earlier version of this work appeared as the preprint ``Do Multilingual VLMs Reason Equally? A Cross-Lingual Visual Reasoning Audit for Indian Languages'' (arXiv:2603.26742).

\bibliography{references}

\appendix
\section{Prompts}\label{app:prompts}
Standard answer instruction (English original): ``Answer with the option letter only (A, B, C, or D) for multiple-choice questions, or a number for numerical questions.'' CoT variant: ``Think step by step, then answer with the option letter only (A, B, C, or D) for multiple-choice questions, or a number for numerical questions.'' The six translated versions are in \texttt{scripts/04\_run\_inference\_open.py} of our repository.

\section{Confidence Intervals and Composition}\label{app:ci}
\begin{table*}[t]
\centering
\scriptsize
\setlength{\tabcolsep}{2.5pt}
\caption{Accuracy (\%) with 95\% percentile-bootstrap confidence intervals (2{,}000 resamples of the 980 items, seed 42), as stored in the released \texttt{summary.json}. Extraction failures are scored as incorrect.}
\label{tab:ci}
\begin{tabular}{@{}lccccccc@{}}
\toprule
\textbf{Model} & \textbf{EN} & \textbf{hi} & \textbf{bn} & \textbf{mr} & \textbf{ta} & \textbf{te} & \textbf{kn} \\
\midrule
GPT-4o & 54.5 [51.4, 57.6] & 44.6 [41.5, 47.6] & 44.5 [41.4, 47.6] & 43.2 [40.0, 46.3] & 40.5 [37.5, 43.6] & 43.0 [39.9, 46.0] & 43.7 [40.5, 46.8] \\
Gemma~3-27B & 51.8 [48.9, 55.0] & 42.0 [38.8, 45.2] & 41.9 [38.9, 45.2] & 42.2 [39.0, 45.5] & 41.9 [38.8, 45.0] & 42.6 [39.5, 45.6] & 41.3 [38.3, 44.6] \\
Llama-4-Maverick & 55.4 [52.2, 58.6] & 43.4 [40.3, 46.4] & 42.7 [39.7, 45.8] & 41.9 [38.9, 45.0] & 34.9 [32.0, 38.0] & 38.6 [35.6, 41.6] & 41.6 [38.5, 44.7] \\
Qwen3-VL-30B & 64.9 [61.9, 68.0] & 44.1 [41.1, 47.2] & 43.3 [40.1, 46.3] & 36.1 [33.0, 39.2] & 41.1 [38.1, 44.1] & 36.5 [33.6, 39.6] & 38.7 [35.6, 41.7] \\
Qwen2.5-VL-32B & 50.3 [47.1, 53.5] & 34.1 [31.1, 37.1] & 39.1 [36.1, 42.4] & 35.6 [32.6, 38.7] & 34.2 [31.2, 37.1] & 33.2 [30.2, 36.1] & 32.6 [29.8, 35.4] \\
Qwen2.5-VL-7B & 51.3 [48.2, 54.4] & 34.2 [31.1, 37.4] & 32.7 [29.7, 35.8] & 30.0 [27.0, 33.1] & 28.7 [25.8, 31.5] & 26.9 [24.2, 29.9] & 30.6 [27.9, 33.6] \\
InternVL2.5-8B & 45.6 [42.5, 48.9] & 29.8 [26.8, 32.8] & 24.9 [22.2, 27.6] & 27.9 [25.1, 30.6] & 22.3 [19.7, 25.0] & 22.9 [20.3, 25.4] & 23.6 [21.0, 26.3] \\
Aya-Vision-8B & 42.0 [38.8, 45.0] & 33.6 [30.6, 36.5] & 21.7 [19.2, 24.4] & 25.1 [22.4, 27.8] & 14.9 [12.7, 17.2] & 11.6 [9.7, 13.6] & 14.2 [12.1, 16.4] \\
Q2.5-7B no-image & 35.8 [32.9, 38.9] & 26.0 [23.3, 28.7] & 24.5 [21.7, 27.2] & 25.1 [22.4, 27.9] & 19.0 [16.6, 21.4] & 20.7 [18.1, 23.4] & 23.7 [21.1, 26.3] \\
Q2.5-7B CoT & 49.5 [46.5, 52.8] & 25.0 [22.4, 27.6] & 18.3 [15.8, 20.7] & 33.3 [30.3, 36.3] & 29.7 [26.8, 32.8] & 22.8 [20.2, 25.5] & 19.2 [16.8, 21.7] \\
\bottomrule
\end{tabular}
\end{table*}

\begin{table}[t]
\centering
\small
\setlength{\tabcolsep}{3pt}
\caption{Composition of the 980 English source items. Items are the first eligible ones in each dataset's official order (no random sampling). MCQ/FF = multiple-choice / free-form numeric.}
\label{tab:composition}
\begin{tabular}{@{}p{0.19\columnwidth}p{0.78\columnwidth}@{}}
\toprule
\textbf{Source} & \textbf{Selection and composition} \\
\midrule
MathVista (400) & 195 MCQ, 205 FF. Testmini PIDs 1--466, keeping questions with $\geq$4 purely alphabetic tokens (66 shorter or non-English questions skipped). Tasks: figure QA 107, VQA 83, math word problems 83, textbook QA 74, geometry 53; 27 source datasets. \\
ScienceQA (400) & 400 MCQ. First 400 of the 1{,}209 natural-science items in the image test split (rows 0--652). Topics: physics 133, biology 125, science practices 56, earth science 51, chemistry 32, other 3; grades 2--8. \\
MMMU (180) & 167 MCQ, 13 FF. All validation items of six STEM subjects (30 each): Math, Physics, Chemistry, Architecture \& Engineering, Mechanical Engineering, Computer Science. \\
\bottomrule
\end{tabular}
\end{table}

\begin{table}[h]
\centering
\small
\setlength{\tabcolsep}{2pt}
\caption{The eight VLMs: seven with open weights and one proprietary model (GPT-4o). Served via follows
the released inference code; exact identifiers and decoding settings are in Appendix~\ref{app:inference}.
vLLM = served locally on one AMD Instinct MI300X GPU; $^{*}$FP8 checkpoint. Overall = macro-average accuracy over the seven languages (\%); \textbf{bold} = best.}
\label{tab:models}
\begin{tabular}{@{}llccr@{}}
\toprule
\textbf{Model} & \textbf{Params} & \textbf{Open} & \textbf{Served via} & \textbf{Overall} \\
\midrule
GPT-4o & n/a & no & OpenAI & \textbf{44.8} \\
Qwen3-VL-30B & 30B-A3B & yes & DeepInfra & 43.5 \\
Gemma~3-27B & 27B & yes & DeepInfra & 43.4 \\
Llama-4-Maverick & 17B$\times$128E & yes & DeepInfra$^{*}$ & 42.6 \\
Qwen2.5-VL-32B & 32B & yes & DeepInfra & 37.0 \\
Qwen2.5-VL-7B & 7B & yes & vLLM & 33.5 \\
InternVL2.5-8B & 8B & yes & vLLM & 28.1 \\
Aya-Vision-8B & 8B & yes & vLLM & 23.3 \\
\bottomrule
\end{tabular}
\end{table}
\section{Inference Details}
\label{app:inference}
\begin{table*}[h]
\centering
\small
\setlength{\tabcolsep}{4pt}
\caption{Model identifiers and serving. All runs use temperature 0, one image per prompt, and
the answer instruction written in the prompt language.}
\label{tab:inference}
\begin{tabular}{@{}ll@{}}
\toprule
\textbf{Model} & \textbf{Identifier (serving)} \\
\midrule
GPT-4o & \texttt{gpt-4o} (OpenAI) \\
Qwen3-VL-30B & \texttt{Qwen/Qwen3-VL-30B-A3B-Instruct} (DeepInfra) \\
Gemma~3-27B & \texttt{google/gemma-3-27b-it} (DeepInfra) \\
Llama-4-Maverick & \texttt{meta-llama/Llama-4-Maverick-17B-128E-Instruct-FP8} (DeepInfra) \\
Qwen2.5-VL-32B & \texttt{Qwen/Qwen2.5-VL-32B-Instruct} (DeepInfra) \\
Qwen2.5-VL-7B & \texttt{Qwen/Qwen2.5-VL-7B-Instruct} (vLLM, bf16) \\
InternVL2.5-8B & \texttt{OpenGVLab/InternVL2\_5-8B} (vLLM, bf16) \\
Aya-Vision-8B & \texttt{CohereForAI/aya-vision-8b} (vLLM, bf16) \\
\bottomrule
\end{tabular}
\end{table*}
Locally served models run with vLLM on one AMD Instinct MI300X GPU (192\,GB) in bfloat16 (maximum context 16{,}384 tokens,
prefix caching enabled), and images are downscaled to at most $512{\times}512$ pixels in area (aspect ratio kept).
API models receive the image downscaled to fit a $512{\times}512$ box. The generation limit is 512 new tokens.
The chain-of-thought variant allows 1{,}536 tokens when run with the local vLLM script but keeps 512 in the API script.

\section{Translation Audit}
\label{app:translation}

\paragraph{What was compared.}
The released verification files contain, for each language, 50 items drawn with Python's \texttt{random.sample} under seed 42, with languages processed in the order hi, ta, te, bn, kn, mr.
We reproduced the identical item lists from the released translations.
The 300 samples cover 256 distinct items (133 MathVista, 122 ScienceQA, 45 MMMU).
Only the \emph{question} string was compared; answer options and the answer instruction were not.
The secondary translation came from Gemini~2.0 Flash, prompted with ``Translate the following English text to \{language\}. Keep numbers, math symbols, and option labels (A, B, C, D) unchanged. Return ONLY the translation, nothing else.'' at the API's default decoding settings.
No human annotator took part in the check.
All 300 IndicTrans2 strings in the verification files are identical to the released translations, so the check was computed on the translations the models saw.
The released repository also contains per-language \texttt{retranslate\_ids.json} lists: 20 distinct items (7 MathVista, 13 MMMU), or 82 item--language pairs (hi 16, ta 17, te 20, bn 9, kn 12, mr 8).
These lists record items that received a second translation pass.
The released text for these pairs still shows placeholder remnants more often than other MathVista/MMMU pairs (28.0\% vs.\ 9.7\%).

\paragraph{Why the reported agreement is weak evidence.}
The score in Table~\ref{tab:languages} is $|\mathrm{set}(a)\cap\mathrm{set}(b)|\,/\,|\mathrm{set}(a)\cup\mathrm{set}(b)|$ over Unicode characters.
It is invariant to word order and repetition: the degenerate Telugu output for mmmu\_math\_0007, which drops the question and ends in a run of 139 ``\$'' characters, still scores 0.78.
Table~\ref{tab:app-null} compares matched pairs with all $50\times49$ mismatched pairs per language (IndicTrans2 output of item $i$ vs.\ Gemini output of item $j\neq i$).
On average the character Jaccard separates matched from unrelated pairs (AUC .91--.99), but at the item level 0.65 lies near the 95th percentile of the unrelated-pair distribution.
It rejects 6--22\% of genuine pairs and accepts 2--17\% of unrelated pairs (4--27\% of length-matched unrelated pairs).
Averaged over 50 items, it is therefore a coarse screen.
The index also cannot tell Hindi from Marathi: the Gemini translation into one language and the IndicTrans2 translation of the same item into the other score 0.57 on average, and 20--32\% of these pairs exceed 0.65.
Sentence-level chrF~\cite{popovic-2015-chrf} and chrF++~\cite{popovic-2017-chrf} (sacreBLEU~\cite{post-2018-call} 2.6.0, default settings) separate the two distributions better.
Even so, 37 of the 57 samples that the triage below marks as major errors or omissions have Jaccard $\geq$0.65.

\begin{table*}[t]
\centering
\footnotesize
\setlength{\tabcolsep}{4pt}
\caption{Agreement between IndicTrans2 and Gemini~2.0 Flash on the 50 verification samples per language, against a null of all 2,450 mismatched item pairs. \emph{Len.-m.}: mismatched pairs whose lengths differ by at most a factor of 1.25. AUC: probability that a matched pair outscores a mismatched one. Takeaway: both scores detect gross mismatch, but the 0.65 Jaccard cut-off accepts many unrelated pairs and rejects many genuine ones.}
\label{tab:app-null}
\begin{tabular}{@{}l ccccc c cccc c@{}}
\toprule
& \multicolumn{5}{c}{\textbf{Set-based char.\ Jaccard}} && \multicolumn{4}{c}{\textbf{chrF}} & \textbf{chrF++} \\
\cmidrule{2-6}\cmidrule{8-11}
& matched & null & \% matched & \% null (len.-m.) & AUC && matched (median) & null & null p95 & AUC & matched \\
& mean & mean & $\geq$.65 & $\geq$.65 & && mean & mean & & & mean \\
\midrule
hi & .844 & .446 & 94 & 4.5 (14.8) & .989 && 68.1 (70.0) & 11.9 & 19.0 & .995 & 67.1 \\
bn & .806 & .446 & 88 & 2.1 (6.3) & .989 && 62.9 (63.0) & 11.7 & 20.4 & .997 & 60.6 \\
mr & .791 & .457 & 78 & 3.8 (8.0) & .948 && 59.2 (61.6) & 11.8 & 21.2 & .959 & 56.8 \\
ta & .811 & .524 & 84 & 17.4 (27.1) & .911 && 65.1 (67.7) & 16.2 & 28.4 & .978 & 60.6 \\
te & .803 & .445 & 88 & 3.3 (8.6) & .962 && 62.6 (63.1) & 12.0 & 20.3 & .985 & 59.3 \\
kn & .791 & .450 & 86 & 2.3 (4.5) & .964 && 60.1 (58.1) & 12.9 & 24.0 & .980 & 55.2 \\
\bottomrule
\end{tabular}
\end{table*}

\paragraph{Automated fidelity checks (all 980 items $\times$ 6 languages).}
We compared every translated item with its English source (Table~\ref{tab:app-fidelity}).
Numbers were compared as multisets after mapping Indic digits to ASCII and removing thousands separators.
The checks also cover leftover placeholder remnants (\texttt{@}, \texttt{<PH>}, \texttt{<ID>}), LaTeX commands and \$ delimiters, strict math symbols, blanks, dropped question marks, length-ratio outliers (below half the language's median ratio), untranslated outputs, answer-option count and structure, and stray unit tokens.
The flags are high-precision but incomplete; for example, placeholders transliterated into the target script are missed.
Three findings stand out.
(i)~Flags are concentrated in MMMU and in long MathVista physics items: the question-level severe flag fires for 63.7\% of MMMU item--language pairs, 8.8\% of MathVista pairs, and 0.1\% of ScienceQA pairs.
(ii)~Flags are largely shared across languages: 194 items are flagged in all six languages, and per-language rates vary by at most about one point.
Pipeline-level changes therefore affect the English-to-Indian drop more than they can explain between-language differences.
(iii)~Some changes are systematic.
The unit-preserving pattern matched word endings (Ye|s, skatin|g, algorith|ms), so the ``Yes'' option of all 77 yes/no items became, for example, Hindi ``ye~s'' or ``aap~s'' (``you~s'').
In the released translations, 36.5--36.7\% of items have a stray unit token in an option.
The option-label pattern keeps a sentence-initial article ``A'' as a Latin letter.
Point labels and angle symbols are often transliterated (Hindi ``ke.~el.'' for KL; ``e.~en.~\d{t}\={\i}.~s\={\i}.~ef.'' for $\angle$DCF), so the text no longer matches the Latin labels drawn in the fixed image.
MMMU \texttt{<image N>} tags are dropped in all 176 items that contain them.
Two MathVista source items (mathvista\_204, mathvista\_243) are in Persian rather than English, and their translations are unrelated text; we recommend excluding them.
Options were translated for all ScienceQA and MMMU MCQs and for 171--172 of the 195 MathVista MCQs.
In 162--164 MathVista MCQs per language, the English answer string no longer matches any translated option, so correctness must be computed against the English option list.
Of the 980 items, 693 (268 MathVista, 389 ScienceQA, 36 MMMU) carry no question- or option-level flag in any language; we release this clean-subset list for robustness checks.

\begin{table}[t]
\centering
\footnotesize
\setlength{\tabcolsep}{2.5pt}
\caption{Automated fidelity flags, \% of the 980 items per language. \emph{Question}: any of the first seven rows. \emph{Options}: broken MMMU option list, altered ``Yes'' option, or changed option count. Takeaway: about a quarter of items carry a flag, at nearly the same rate in every language.}
\label{tab:app-fidelity}
\begin{tabular}{@{}lcccccc@{}}
\toprule
\textbf{Flag} & \textbf{hi} & \textbf{bn} & \textbf{mr} & \textbf{ta} & \textbf{te} & \textbf{kn} \\
\midrule
Number missing & 7.7 & 6.9 & 6.5 & 9.5 & 7.0 & 7.1 \\
Number added & 9.4 & 10.1 & 10.7 & 8.1 & 10.0 & 8.5 \\
Placeholder remnants & 6.7 & 6.4 & 6.7 & 5.1 & 5.3 & 5.6 \\
LaTeX command lost & 3.0 & 2.5 & 3.2 & 3.0 & 2.6 & 2.9 \\
``?'' dropped & 2.9 & 3.0 & 2.8 & 3.4 & 2.6 & 3.4 \\
Length ratio low & 0.7 & 0.6 & 0.5 & 1.5 & 0.9 & 1.1 \\
Untranslated & 0.4 & 0.4 & 0.3 & 0.1 & 0.4 & 0.3 \\
\midrule
Math symbol lost & 6.2 & 6.0 & 6.1 & 6.9 & 6.0 & 6.6 \\
Latin label lost & 10.8 & 11.2 & 13.0 & 10.4 & 6.9 & 11.4 \\
MMMU options broken & 3.8 & 3.9 & 3.7 & 3.7 & 3.5 & 3.7 \\
``Yes'' option altered & 7.9 & 7.9 & 7.9 & 7.9 & 7.9 & 7.9 \\
\midrule
\textbf{Question} & 15.6 & 15.0 & 15.1 & 16.0 & 15.3 & 15.1 \\
\textbf{Options} & 11.6 & 11.7 & 11.5 & 11.6 & 11.3 & 11.5 \\
\textbf{Either} & 24.2 & 23.8 & 23.6 & 24.5 & 23.8 & 23.6 \\
\bottomrule
\end{tabular}
\end{table}

\paragraph{LLM-assisted adequacy triage (not human-validated).}
A large language model read the English source, the IndicTrans2 question, and the Gemini output for all 300 verification samples, using the Gemini output as a second reference.
It labelled each question translation as \emph{adequate}, \emph{minor issue} (calque or non-standard term, untranslated fragment, stray token, transliterated label), \emph{major meaning error} (changed value, relation, quantifier, or unit; wrong-concept term; lost notation), or \emph{omission}.
Answer options were excluded here because the automated checks cover them.
These labels are a triage aid only and must be confirmed by native speakers.
Table~\ref{tab:app-triage} summarises them and Table~\ref{tab:app-examples} lists verifiable examples.
Major errors and omissions concentrate in MMMU (46.7\%) and MathVista (22.6\%), against 4.9\% in ScienceQA.
The most frequent omission is the \emph{final sentence}, often the actual question, of multi-sentence items; each item was passed to the model as a single unsplit input.
The Gemini reference itself had issues in 14 of the 300 samples. Six are substantial: answering or summarising instead of translating, translating only a fragment, or mixing in Arabic words. Eight are minor, e.g., dropped instruction sentences, stray Hebrew or Chinese tokens, or genus rendered as `species'.
Low agreement therefore does not always indicate an IndicTrans2 error.

\begin{table}[t]
\centering
\footnotesize
\setlength{\tabcolsep}{3pt}
\caption{LLM-assisted triage of the 300 verification samples (question text only): adequate (A), minor issue (MI), major meaning error (MA), omission/truncation (TR). CI: Wilson 95\%. Takeaway: roughly one sample in five has an answer-relevant error, mostly in MMMU and MathVista; labels await native-speaker confirmation.}
\label{tab:app-triage}
\begin{tabular}{@{}lrrrrc@{}}
\toprule
& \textbf{A} & \textbf{MI} & \textbf{MA} & \textbf{TR} & \textbf{MA+TR \% [CI]} \\
\midrule
hi & 33 & 11 & 4 & 2 & 12.0 [5.6, 23.8] \\
bn & 25 & 14 & 9 & 2 & 22.0 [12.8, 35.2] \\
mr & 25 & 18 & 4 & 3 & 14.0 [7.0, 26.2] \\
ta & 25 & 16 & 5 & 4 & 18.0 [9.8, 30.8] \\
te & 24 & 17 & 5 & 4 & 18.0 [9.8, 30.8] \\
kn & 21 & 14 & 5 & 10 & 30.0 [19.1, 43.8] \\
\midrule
All (300) & 153 & 90 & 32 & 25 & 19.0 [15.0, 23.8] \\
MathVista (133) & 71 & 32 & 18 & 12 & 22.6 [16.3, 30.4] \\
ScienceQA (122) & 73 & 43 & 3 & 3 & 4.9 [2.3, 10.3] \\
MMMU (45) & 9 & 15 & 11 & 10 & 46.7 [32.9, 60.9] \\
\bottomrule
\end{tabular}
\end{table}

\begin{table*}[t]
\centering
\footnotesize
\setlength{\tabcolsep}{3pt}
\caption{Examples of translation issues (English glosses of the released IndicTrans2 output). Takeaway: the issues that change answers are omissions of the final question, altered quantities, and wrong-concept terms, not fluency.}
\label{tab:app-examples}
\begin{tabular}{@{}llp{0.33\textwidth}p{0.40\textwidth}@{}}
\toprule
\textbf{Item} & \textbf{Lang.} & \textbf{English source (excerpt)} & \textbf{IndicTrans2 output, glossed} \\
\midrule
mathvista\_430 & ta & The derivative of f(x) at x=0 is \_\_\_\_ that at x=5 & ``the derivative of f(x), x=0 is x=5'' (blank and comparison dropped) \\
mathvista\_54 & ta & \dots\ is 2 : 5. If AC = 25, what is the length of line AB? & first sentence only; the question is dropped \\
mathvista\_331 & hi & \dots\ What is the mean of the numbers? & ``What is the \emph{meaning} of the numbers?'' \\
mathvista\_455 & bn & \dots\ what is the length of the arc shown \dots & ``\dots\ length of the \emph{circle} \dots'' \\
mmmu\_physics\_0022 & ta & \dots\ resistance $10^{-2}$ ohm \dots\ 0.4 tesla \dots & ``resistance 10 ohm''; 0.4 tesla dropped \\
mmmu\_physics\_0021 & hi & \dots\ from the n=3 to the n=1 level instead \dots & ``from n=6~@ to n=8~@ level'' (placeholder shift) \\
mathvista\_186 & bn & \dots\ AC = 3.0, BD = 6.0, CD = 10.0 \dots & ``AC = 5~@, BD = <@50>~@, CD = 10.0'' \\
mathvista\_392 & hi & How many people lifted at least 46 pounds? & number replaced by transliterated tags ``P.~H.~2'' \\
mathvista\_373 & mr & As shown in the figure, in the quadrilateral ABCD, $\angle$BAD = 120.0 \dots & repetition loop ``A.~B.~C.~D., A.~B.~C.~D., \dots\ D.~D.~D.'' \\
mathvista\_404 & hi & options: figure skating practice; private class \dots & ``figure skatin~g practice'', ``private class~s'' \\
77 yes/no items & all & options: Yes / No & ``Yes'' $\rightarrow$ ``ye~s''/``you~s'' (e.g., Hindi ``aap~s'') \\
mathvista\_204 & all & (source is Persian, not English) & unrelated text, e.g., Marathi ``today's city runs for two days'' \\
\bottomrule
\end{tabular}
\end{table*}

\paragraph{Native-speaker validation protocol.}
We release the following protocol for native-speaker validation as a resource for follow-up work.
\emph{Sample:} 100 items per language (600 in total): the 50 already-triaged items, so the LLM triage can be scored against humans, plus 50 new items drawn with a fixed seed, with MMMU oversampled (20 of the 50) because it carries most defects.
Prevalence estimates are reweighted to the 400/400/180 source mix.
\emph{Annotators:} two per language, native speakers who studied or teach STEM in that language medium.
They work independently and are blind to model outputs, automated flags, and LLM labels.
\emph{Inputs:} the image, the English question and options, and the translated question and options.
\emph{Labels:} (1)~the four-way adequacy label defined above; (2)~a binary \emph{answer-preserving} judgement (``could a competent student reading only the translation and the image reach the same correct answer?''); (3)~whether the options are faithful; and (4)~for major errors and omissions, a corrected translation, which enables a post-edited v2 of the benchmark.
\emph{Reporting:} per-language Cohen's $\kappa$ and raw agreement for labels (1) and (2); adjudication of disagreements by a third annotator; prevalence of non-answer-preserving items with Wilson 95\% CIs (13--29\% for an observed 20\% at $n{=}100$); $\kappa$ between adjudicated human labels and the LLM triage; and the precision and recall of the automated flags.
Accuracy drops are then re-computed on human-validated clean items, which requires per-item model outputs (script released).
At roughly 1--3 minutes per judgement, the 1,200 judgements need about 20--60 annotator-hours.

\section{Text inside benchmark images}
\label{app:imagetext}
We ran OCR (RapidOCR, PP-OCRv3 models) on all 980 images at full resolution and classified each item as:
(a) no text;
(b) only language-neutral tokens (numbers, single letters or variable names, units, chemical formulas, short all-caps labels);
(c) English words, meaning at least one real English word of three or more letters, such as an axis title, legend or label;
(d) text-heavy, meaning a table (at least three rows in which an English word cell is followed by a numeric cell), at least two sentence-length lines outside the title band, or at least 40 English words.
Table~\ref{tab:imagetext} gives the distribution.
To check the rules, we visually inspected a stratified random sample of 40 images (seed 2026), stratified by source and OCR category.
The binary split that matters here, English words present (c+d) versus absent (a+b), matched in 39/40.
When images are downscaled to at most 512\,px, as in our inference scripts, OCR still finds English words in 422 of the 423 (c)+(d) images.
In 29.7\% of items, a content word of the English question or options also appears as an English word in the image.

\begin{table}[H]
\centering
\small
\setlength{\tabcolsep}{3pt}
\caption{English text inside the fixed images (OCR-based, \% of items). English = (c)+(d). Takeaway: over half of items have language-neutral images; 43\% carry English words, mostly short labels rather than running text.}
\label{tab:imagetext}
\resizebox{\columnwidth}{!}{%
\begin{tabular}{@{}lrrrrr@{}}
\toprule
Source (n) & (a) none & (b) neutral & (c) words & (d) heavy & English \\
\midrule
MathVista (400) & 37.2 & 16.0 & 38.5 & 8.2 & 46.8 \\
ScienceQA (400) & 57.8 & 2.8 & 36.2 & 3.2 & 39.5 \\
MMMU (180)      & 14.4 & 42.2 & 41.1 & 2.2 & 43.3 \\
\midrule
All (980)       & 41.4 & 15.4 & 38.1 & 5.1 & 43.2 \\
\bottomrule
\end{tabular}%
}
\end{table}

\section{Answer Extraction: Failure Rates and Sensitivity}
\label{app:extraction}
Failure rates per model and language are in Table~\ref{tab:failure-rates}; the sensitivity scenarios are in Table~\ref{tab:extraction-sensitivity}.
\begin{table*}[t]
\centering
\small
\setlength{\tabcolsep}{4pt}
\caption{Extraction-failure rate (\% of the 980 items with no extractable letter or number) per model and language. IA/DR = pooled over the three Indo-Aryan/Dravidian languages. \textbf{Takeaway:} failures are concentrated in a few models and, for Aya-Vision-8B, overwhelmingly in Dravidian languages (54.3\% vs.\ 21.9\%); for GPT-4o and Gemma~3-27B they are low and flat across languages.}
\label{tab:failure-rates}
\begin{tabular}{@{}l r ccc ccc cc@{}}
\toprule
\textbf{Model} & \textbf{EN} & \textbf{hi} & \textbf{mr} & \textbf{bn} & \textbf{ta} & \textbf{te} & \textbf{kn} & \textbf{IA} & \textbf{DR} \\
\midrule
Qwen3-VL-30B & 3.4 & 10.4 & 11.0 & 11.8 & 11.6 & 21.0 & 18.5 & 11.1 & 17.0 \\
GPT-4o & 5.7 & 6.6 & 6.5 & 7.7 & 7.0 & 8.1 & 7.4 & 6.9 & 7.5 \\
Gemma~3-27B & 6.3 & 6.6 & 4.8 & 6.7 & 5.4 & 6.0 & 5.9 & 6.1 & 5.8 \\
Llama-4-Maverick & 2.8 & 5.4 & 5.5 & 6.5 & 14.1 & 7.7 & 7.6 & 5.8 & 9.8 \\
Qwen2.5-VL-32B & 21.3 & 25.0 & 15.0 & 16.7 & 12.1 & 25.3 & 13.6 & 18.9 & 17.0 \\
Qwen2.5-VL-7B & 6.6 & 10.6 & 20.8 & 20.4 & 23.1 & 27.0 & 16.5 & 17.3 & 22.2 \\
InternVL2.5-8B & 16.2 & 17.8 & 23.2 & 29.5 & 35.2 & 24.7 & 28.8 & 23.5 & 29.6 \\
Aya-Vision-8B & 9.1 & 15.5 & 19.3 & 30.8 & 50.9 & 58.9 & 53.0 & 21.9 & 54.3 \\
\midrule
\multicolumn{10}{@{}l}{\textit{Qwen2.5-VL-7B ablations}} \\
No-image & 24.8 & 24.5 & 19.5 & 26.7 & 27.2 & 32.1 & 19.1 & 23.6 & 26.2 \\
Chain-of-thought & 0.5 & 23.6 & 17.1 & 33.8 & 17.3 & 28.7 & 33.6 & 24.8 & 26.5 \\
\bottomrule
\end{tabular}
\end{table*}

\begin{table*}[t]
\centering
\small
\setlength{\tabcolsep}{4pt}
\caption{Extraction-failure sensitivity of the Dravidian$-$Indo-Aryan gap. Fail = share of the 980 items with no extractable answer, pooled over the three Indo-Aryan (IA) or Dravidian (DR) languages. Gap = Dravidian minus Indo-Aryan accuracy drop (pp) under S1 (strict: failures wrong; Table~\ref{tab:family}), S2 (accuracy among validly extracted responses only), S3 (failures credited at the nominal chance rate $p=0.284$), and S5 (every failure credited as correct; an upper bound, not an estimate). $r^\ast$ = hidden accuracy among failures, common to both families, that would erase the S1 gap (``none'': Dravidian conditions have fewer failures, so crediting failures widens the gap). \textbf{Takeaway:} the gap stays positive for 7/8 models under S2 and 8/8 under S3 but shrinks to at most 4.5$\pp$ (S2); about two-thirds of Aya-Vision-8B's 13.2$\pp$ gap depends on how its Dravidian-skewed extraction failures are scored.}
\label{tab:extraction-sensitivity}
\begin{tabular}{@{}l rr rrrr r@{}}
\toprule
 & \multicolumn{2}{c}{\textbf{Fail (\%)}} & \multicolumn{4}{c}{\textbf{Gap (pp)}} & \\
\cmidrule(lr){2-3}\cmidrule(lr){4-7}
\textbf{Model} & \textbf{IA} & \textbf{DR} & \textbf{S1} & \textbf{S2} & \textbf{S3} & \textbf{S5} & $\boldsymbol{r^\ast}$ \\
\midrule
Qwen3-VL-30B & 11.1 & 17.0 & +2.4 & $-$0.5 & +0.7 & $-$3.6 & 0.40 \\
GPT-4o & 6.9 & 7.5 & +1.7 & +1.5 & +1.5 & +1.1 & 2.94 \\
Gemma~3-27B & 6.1 & 5.8 & +0.1 & +0.3 & +0.2 & +0.4 & none \\
Llama-4-Maverick & 5.8 & 9.8 & +4.3 & +2.8 & +3.2 & +0.3 & 1.09 \\
Qwen2.5-VL-32B & 18.9 & 17.0 & +3.0 & +4.4 & +3.5 & +4.9 & none \\
Qwen2.5-VL-7B & 17.3 & 22.2 & +3.5 & +2.1 & +2.1 & $-$1.4 & 0.72 \\
InternVL2.5-8B & 23.5 & 29.6 & +4.6 & +3.3 & +2.9 & $-$1.5 & 0.75 \\
Aya-Vision-8B & 21.9 & 54.3 & +13.2 & +4.5 & +4.1 & $-$19.1 & 0.41 \\
\midrule
\multicolumn{8}{@{}l}{\textit{Qwen2.5-VL-7B ablations}} \\
No-image & 23.6 & 26.2 & +4.1 & +4.4 & +3.3 & +1.5 & 1.58 \\
Chain-of-thought & 24.8 & 26.5 & +1.6 & +1.2 & +1.2 & $-$0.1 & 0.96 \\
\bottomrule
\end{tabular}
\end{table*}

\end{document}